\documentclass[lettersize,journal]{IEEEtran}
\IEEEoverridecommandlockouts
\usepackage{enumitem}
\usepackage{graphicx}
\usepackage{adjustbox}
\usepackage{cite}
\usepackage{amsmath,amssymb,amsfonts}
\usepackage{algorithmic}
\usepackage[caption=false,font=normalsize,labelfont=sf,textfont=sf]{subfig}
\usepackage{textcomp}
\usepackage{xcolor}
\usepackage{wrapfig}
\usepackage{float}
\usepackage{hyperref}
\usepackage{enumitem}

\usepackage[commentmarkup=uwave]{changes}
\definechangesauthor[name={Santiago Lopez}, color=red]{slt}
\def\bc{{\mathbf{c}}}

\def\bn{{\mathbf{n}}}

\def\bx{{\mathbf{x}}}
\def\by{{\mathbf{y}}}
\def\bz{{\mathbf{z}}}
\def\b0{{\mathbf{0}}}

\def\bH{{\mathbf{H}}}

\begin{document}

\title{VDPI: Video Deblurring with Pseudo-inverse Modeling\\
}

\author{
\IEEEauthorblockN{Zhihao Huang, Santiago López-Tapia and Aggelos K. Katsaggelos, \textit{Fellow, IEEE}
\thanks{Zhihao Huang, Santiago López-Tapia and Aggelos K. Katsaggelos are with the McCORMICK SCHOOL OF ENGINEERING, Northwestern University, Evanston 60201, Illinois  (e-mail:\href{zhihaohuang2022@u.northwestern.edu}{zhihaohuang2022@u.northwestern.edu}). }% <-this % stops a space
\thanks{}
\IEEEpubid{0000--0000/00\$00.00~\copyright~2021 IEEE}
}

}
% \and
% \and
% \and
% \and

\maketitle

\begin{abstract}

Video deblurring is a challenging task that aims to recover sharp sequences from blur and noisy observations. The image-formation model plays a crucial role in traditional model-based methods, constraining the possible solutions. However, this is only the case for some deep learning-based methods. Despite deep-learning models achieving better results,  traditional model-based methods remain widely popular due to their flexibility. An increasing number of scholars combine the two to achieve better deblurring performance. This paper proposes introducing knowledge of the image-formation model into a deep learning network by using the pseudo-inverse of the blur. We use a deep network to fit the blurring and estimate pseudo-inverse. Then, we use this estimation, combined with a variational deep-learning network, to deblur the video sequence. Notably, our experimental results demonstrate that such modifications can significantly improve the performance of deep learning models for video deblurring. Furthermore, our experiments on different datasets achieved notable performance improvements, proving that our proposed method can generalize to different scenarios and cameras.

\end{abstract}


\section{Introduction}
Videos captured by hand-held cameras in dynamic environments often suffer from various levels of blur\cite{b1}\cite{b2} caused by object motion or camera shake\cite{b3}. This problem has received significant attention as the blur in the videos usually interfere with subsequent high-level vision tasks. Video deblurring aims to restore a clear sequence from the blur and noisy observation, making it a critical task in video processing. Mathematically, the blurred image $\by$ is modeled as a convolution of the latent image $\bx$ and the blur $\bH$ and the addition of noise $\bn$ as: 
    \begin{equation}
    \by = \bH\bx + \bn.
    \label{eq:im_for1}
    \end{equation}
We consider $\bn$ to be, as in most applications, Additive White Gaussain Noise (AWGN). To render this problem more tractable, conventional approaches typically develop different image priors, which constitute the first category: model-based methods\cite{b4}. Model-based approaches explicitly define and utilize the degradation process described by the image formation model in \eqref{eq:im_for1} in an optimization procedure.\cite{katsaggelos2007super}\cite{belekos2010maximum}. Most model-based approaches make use of hand-crafted priors or regularization to address the ill-posedness of the problem.\cite{lopez-tapia:DSP:2020}. Early studies on video deblurring largely focused on removing uniform\cite{b8} and non-uniform blurs\cite{b9,b10} by aggregating multiple images. To tackle complex spatially-varying motion blur, Wulff and Black\cite{b11} pioneered a novel layered model that estimated layer segmentation and restored foreground and background blurry regions separately. Kim and Lee\cite{b12} approximated pixel-wise kernels using bidirectional optical flows as a solution to the deblurring problem. The approach proposed in \cite{b8} introduced a unified multi-image blind deconvolution algorithm to recover clean images from various degraded, blurry inputs. Meanwhile, \cite{b9} extracted blur kernels and mitigated them by computing the duty cycle of the video sequence. However, a common limitation of these model-based methods lies in the need to design intricate energy functions, which are arduous and time-consuming to optimize.

To circumvent the challenges faced by conventional methods, deep learning-based approaches have emerged as a promising alternative \cite{b13,b14,b15,b16,b17,b18,b19}. In contrast to model-based techniques that explicitly define and utilize image formation models, learning-based methods leverage large training databases to directly restore clear images/videos from their blurred counterparts, removing the need for handcrafted priors. One such approach by Su et al.\cite{b20} introduced a simple convolutional neural network that takes five consecutive frames as input and restores the middle frame. To effectively exploit information across multiple frames, their method employed homography alignment and optical flow alignment techniques capable of handling severe blur. The core strength of learning-based deblurring methods lies in their data-driven nature, bypassing the explicit modeling of the degradation process and associated ill-posed inverse problems that plague conventional optimization-based approaches. Moreover, Tao et al. introduced the SRN \cite{b13}, which employs a scale-recurrent architecture to progressively restore high-resolution details in blurry videos by leveraging the hierarchical structure of video data. Similarly, Zhang et al. proposed the STRCNN \cite{b18}, utilizing a recurrent framework to handle spatially varying blur, enhancing the clarity of dynamic scenes. Building on these approaches, Tao et al. developed the DBN \cite{b15}, which exploits both spatial and temporal information to effectively reduce blur in videos, demonstrating significant improvements in handling complex motion blur. The EDVR method \cite{b24} utilizes deformable convolutions to align and fuse video frames, significantly improving deblurring performance by adapting to the motion and structure of the scene. Son et al. introduced the PVDNet \cite{Son2021PVDNet}, a network that applies a coarse-to-fine strategy to progressively refine video frames for superior deblurring results. Pan et al. presented the CDVD-TSP \cite{b26}, which employs a temporal sharpness prior and non-local spatial-temporal similarity to guide the deblurring process, resulting in sharper and more coherent video frames. The IFI-RNN \cite{IFI-RNN} iteratively infers intermediate latent frames, leveraging temporal information to enhance deblurring performance. In the same vein, Pan et al. proposed the TSP method \cite{b28}, which focuses on utilizing sharpness priors from adjacent frames to guide the deblurring process, effectively handling significant blur effects in videos. More recently, the SAPHN\cite{SAPHN} incorporates spatial attention mechanisms to focus on crucial areas within video frames, improving deblurring performance by prioritizing important regions. Finally, the ESTRNN \cite{ESTRNN} combines spatio-temporal information with recurrent networks to achieve efficient and high-quality deblurring results.

Although convolutional neural network (CNN) based models typically outperform their model-based counterparts, most of them lack the flexibility inherent to model-based methods. This reduced flexibility stems from the fact that CNN models are trained to handle a specific type of degradation operator. Consequently, the trained model is limited to addressing only one type of degradation, and its performance deteriorates significantly when a mismatch occurs between the degradation models used during training and testing phases\cite{b31}\cite{b32}. To mitigate the aforementioned limitations, model-based approaches appear to offer a potential avenue for improvement. Consequently, researchers have attempted to combine model-based and learning-based approaches. In \cite{b33}, the author illustrated a solution. They used a Wiener filter to approximate the Moore-Penrose pseudo-inverse of the blur convolution operator. The problem solve by the CNN is then reformulated as learning a residual in the null space of the blur kernel, going from a deconvolution problem to a denoising one. This residual, when added to the Wiener restoration, satisfies the image formation model. This approach is advantageous because the network needs to learn the residuals associated with the Wiener filter, separating its task from the blur on the image, thus making it capable of handling various blurs effectively. However, this conecpt cannot be easily applied to non-uniform motion blur, where the blur is unown and changes per pixel. However, based on the effectiveness and versatility of CNNs, we can consider using them to approximate both this unknown non-uniform blur and its the pseudo-inverse. Based on this concept, it is natural to consider using the effectiveness and versatility of CNNs to achieve the fitting of the blur kernel and the pseudo-inverse kernel. Using the estimation of the pseudo-inverse, we can provide CNNs for non-uniform motion blur of the benefits shown in \cite{b33}, allowing CNNs to flexibly adapt to and fit various scenarios, thereby improving the effectiveness and accuracy of the deconvolution process.

Therefore, our work is divided into three steps:
\begin{itemize}
\item[\large\textbullet]
\textit{Using CNNs to fit the blurring process $\bH$}
\item[\large\textbullet]
\textit{Based on this $\bH$, also using CNNs we estimate the pseudo-inverse $\bH^{+}\by$.}
\item[\large\textbullet]
\textit{Introducing the estimated $\bH^{+}\by$ into the variational deep-learning network to obtain the sharp ones.}
\end{itemize}

\begin{figure}[!htb]
    \centering
    \includegraphics[width=0.49\textwidth]{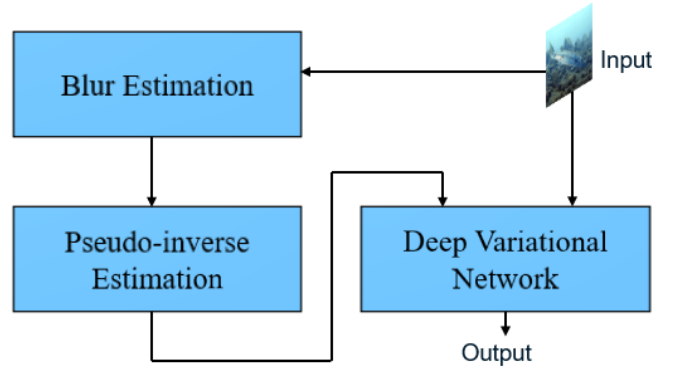}
    \caption{The proposed network framework consists of three main components: Blur Estimation, Pseudo-inverse Estimation, and Deep Variational Network.}
    \label{fig1}
\end{figure}

Fig. \ref{fig1} illustrates the overall structure and process of the network. The process begins with the input image being fed into the Blur Estimation module, where the extent and nature of the blur are assessed. Subsequently, this estimated blur information is passed to the Pseudo-inverse Estimation module, which calculates a pseudo-inverse operator to aid in the deblurring process. Both the output from the Pseudo-inverse Estimation and the original input image are then provided as inputs to the Deep Variational Model. This model integrates the pseudo-inverse operator and the original image data to perform comprehensive deblurring, ultimately producing the final deblurred output image.

The rest of the paper is structure as follows: Section\ref{sec:method}, we present our model and the network structure. Section\ref{thrd:setting} presents details on the experimental setup and training process. We perform an ablation study to show the contributions of each proposed component in Section \ref{four:ablation} using the Gopro \cite{b38}, DVD \cite{b39}, and REDS \cite{Nah_2019_CVPR_Workshops_REDS} datasets. In Section\ref{five:comparison}, we perform a performance comparison with the state-of-art on the same datasets. Finally, conclusions are presented in Section\ref{six:conclusion}.

\section{Methodology}
\label{sec:method}
As previously stated, video deblurring is a very challenging task for to the variability of the degradations and the ill-posed nature of the problem. Even when considering the best possible scenario, known blur and no noise ($\bn = 0$ in \eqref{eq:im_for1}), the estimation of the sharp image is not a trivial task. From \eqref{eq:im_for1}, to calculate the sharp image $\bx$ in this ideal scenario, we should:
    \begin{equation}
    \bx = \bH^{-1}\by,
    \label{eq:im_for2}
    \end{equation}
However, the inverse of $\bH$ may not exist. This problem can be tackle by using the pseudo-inverse of the blur, $\bH^{+}$. It can be shown\cite{b34} that:
    \begin{equation}
    \bH^{+} = \lim_{{\delta \to 0^+}}(\bH^{T}\bH+\delta I)^{-1}\bH^T,
    \label{eq:im_for3}
    \end{equation}
Therefore, $\bH^{+}\by$ retrieves the frequencies of $\bx$, except for those for which the Fourier transform of the blurring filter is zero. These frequencies constitute the null space of the kernel $\bH$\cite{meyer_matrix_2010}. It's important to note that for these frequencies, reversing the blur effects using the pseudo-inverse is not possible, leading to the emergence of artifacts. Consequently, there arises a necessity to recover (learn) those frequencies from the original image. The remaining frequencies are contained within $\bH^{+}\by$. Reasonably, $\bH^{+}\by$ is a good estimation which can be the compensation of restoration task. 

The core operation of the model-based approaches mentioned above is the definition and computation of $\bH$. The task of deblurring when the blur process $\bH$ is known is referred to as Non-Blind Image Deconvolution (NBID) \cite{meyer_matrix_2010}. To enrich the variety of blur kernels while avoiding the tedious and complex definition and computation, some methods use the recurrence of patches within the image and across scales, such as those proposed in\cite{b36}\cite{b37}. While these approaches perform well on repetitive structures, they do not achieve state-of-the-art results on natural images \cite{b37}. Also, one limitation is that these methods assume the blur kernel $\bH$ to be known and consistent across all training and testing images. Consequently, these methods are designed for a single blur kernel only, such as bicubic or an isotropic Gaussian with a fixed kernel width.
Therefore, these methods do not solve the problem well, because it is difficult to categorize images and videos in large modern datasets by specific types of blur, as they are often mixed together. According to our idea, using CNNs to learn the blur process $\bH$ is an excellent solution, as neural networks are inherently well-suited for fitting such complex features and parameters.

\subsection{\textit{Bluring Simulation with the blurred one $\by$ only}}
Recently, Chu et al. \cite{chu2022nafssr} proposed a network block called NAFNet. This block reduces computational complexity at both the intro-block and inner-block levels,
    \begin{figure*}[h!]
      \centering
      \includegraphics[width=1\linewidth]{Cal_H.pdf}
      \caption{The network to simulate blurring kernels $\bH$, which consisted of NAFNet Block, Down-Sample Block, Up-sample Block. Particularly, We use NAFNet as both the encoder and decoder. At each upsampling level, the network outputs the corresponding blurring simulations' features $\bH_0$, $\bH_1$, $\bH_2$ (with shape 32, 64 and 128). It is important to note that the network's input is only the blurred image y. While we often have sharp images as ground truth during training, in some test sets, we only have the blurred images. Therefore, we combine the ground truth to calculate the loss during training, but the network's input does not require the sharp image x. The intuitive understanding is that once the network is trained, the input can be any image, whether sharp or blurred.
    }
      \label{fig2}
    \end{figure*}
while also delivering excellent performance on restoration tasks. Considering our objective involves additional network training, we chose the NAFNet block as our feature extraction unit.
The overarching goal is to achieve a meticulously trained $\bH$, along with its corresponding product with vector y, denoted as $\bH^+\by$. To accomplish this objective, I plan to break down the entire process into two distinct steps: first, the computation of $\bH$, followed by its application to the vector y. The network architecture employed in the initial step is meticulously depicted in Fig. \ref{fig2}. Initially, the blurred input y is concatenated along the channel dimension to form a single input variable for the network. Feature extraction is performed by a 3x3 CNN, followed by refinement through the NAFNet block. These refined features then pass through a downsampler, consisting of 2x2 CNNs, which iteratively condense them into varying-dimensional representations.

After obtaining the features representing the blurring process, the next step is to determine how to get $\bH\bx$ or $\bH\by$. Unlike analytical methods, we still use a learning-based approach for this step. Additionally, this second step is where we introduce the training loss for the blurring simulation. From \eqref{eq:im_for1}, we can clearly see that the difference between y and $\bH\bx$ can serve as a suitable Charbonnier loss function:
\begin{equation}
    Blur\_Loss = \sum\limits_{i=0}^{2}\| \by_i - \bH\bx_i \|^2_2
    \label{eq:im_for4}
\end{equation}

Next, I will introduce the BlurDictModel, which leverages the network $\bH$. As illustrated in Fig. \ref{fig3}, this model integrates $\bH$ with the input in a structured manner. Initially, we perform mean subtraction on both the first and second dimensions of the input to normalize the data. The resulting features are then fed into the ReplicationPad layer, with a padding length set to 7, which is half the size of the CNN kernel.

After padding, the features are processed through two Conv3d layers. These layers are designed to extract features for both the input and $\bH$ channels separately. Each Conv3d layer has a kernel size of $1\times 15\times 15$. The key difference between the two layers lies in their output dimensions: the Conv3d layer processing the $\bH$ channel produces 50 output features, while the layer processing the input channel generates a single output feature.

Once the feature extraction is complete, we apply a similar mean subtraction process to the extracted features. This step ensures consistency and further normalizes the data. Finally, we sum the processed features, which yields the desired outputs such as $\bH\bx$, $\bH\by$, or any other required outputs.

This structured approach allows the BlurDictModel to effectively integrate and process the input and $\bH$ channels, leveraging the strengths of the Conv3d layers and mean normalization to produce accurate and relevant outputs.

As mentioned earlier, to enhance feature richness and performance during training, the feature training network for $\bH$ outputs three different-dimensional features: $\bH_0$, $\bH_1$, $\bH_2$. Consequently, during the $\bH\bx$ (or $\bH\by$) simulation phase, computations for various dimensions are required. This necessitates providing the BlurDictModel with three different-dimensional inputs. To achieve this, we use Laplacian sampling to obtain the corresponding three dimensions of the input. Laplacian sampling allows us to efficiently capture and represent different scales of the input data, providing a robust foundation. 

\begin{figure}[!htb]
    \centering
    \includegraphics[width=0.42\textwidth]{BlurDictModel.pdf}
    \caption{BlurDictModel structure, which consisted of two branches: input branch and $\bH$ branch. The key feature extractors are two Conv3d layers with kernel size $1\times 15\times 15$. One of them will output the feature with channel size 50 to multiply with input feature $\bH$, then merge with the feature from input branch.
    }
    \label{fig3}
\end{figure}

 To elaborate further, the overall structure comprises three components of Laplacian sampling networks connected sequentially, each outputting corresponding-sized high-frequency and low-frequency sampled images. The downsampler consists of a ReplicationPad Layer followed by a CNN Layer. Similarly, the upsampler replicates this structure, but with an additional upsampling operation preceding the padding. Hence, the low-frequency component originates from the downsampling results, while the high-frequency component stems from the difference between the input and the low-frequency components.

\begin{table}[!htb]
\centering
    \caption{Metrics of Blurring Process with only y. The PSNR, which measures image quality, is high, averaging at 38.85. SSIM values are near perfect, with a mean of 0.995.}
    \vskip 10pt
    \begin{tabular}{ccccccc}
    \hline \noalign{\smallskip}
    \multicolumn{1}{c}{} & \multicolumn{1}{c}{Valid1} & \multicolumn{1}{c}{Valid2} & \multicolumn{1}{c}{Valid3} & \multicolumn{1}{c}{Valid4} & \multicolumn{1}{c}{Valid5} & \multicolumn{1}{c}{Mean} \\ \noalign{\smallskip} \hline \noalign{\smallskip}
    Loss & 0.018       & 0.018       & 0.015       & 0.018       & 0.025       & 0.019   \\ \noalign{\smallskip}
    PSNR & 39.05       & 39.53       & 37.92       & 38.54       & 39.21       & 38.85   \\ \noalign{\smallskip}
    SSIM & 0.995       & 0.995       & 0.996       & 0.994       & 0.993       & 0.995   \\ \noalign{\smallskip} \hline
    \end{tabular}
    \label{table1}
\end{table}
    
\begin{figure}[!htb]
    \centering
    \includegraphics[width=0.49\textwidth]{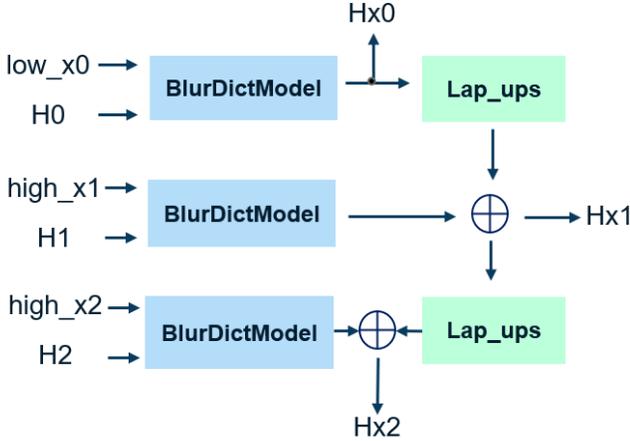}
    \caption{The entire structure of applying blurring kernels to the input. Combining different dimension levels and output three dimension outputs $\bH\bx_0$, $\bH\bx_1$, $\bH\bx_2$. Which could be used for any input, not only $\bx$.}
    \label{fig4}
\end{figure}

Finally, the structure of the entire $\bH\bx$ or $\bH\by$ simulation, which I refer to as the
ApplyH network, is depicted in Fig. \ref{fig4}. At each dimension level, calculating the result by BlurDictModel and summing up with the upsampling from last dimension level. So far, we have introduced the methods for simulating the blurring process. Since this involves the simulation of the blurring process, our main focus is on comparing the PSNR and SSIM between y and $\bH\bx$. We conducted simulations on the Gopro\cite{b38}, DVD\cite{b39}, and REDS\cite{Nah_2019_CVPR_Workshops_REDS} datasets. For brevity, we only present the metrics 
based on the REDS dataset, as shown in Table \ref{table1}. Before comparison, we converted both from RGB space to YCbCr space\cite{b42}.

We conducted five rounds of validation, aggregating the results and deriving the mean as the conclusive metric. The analysis revealed an overall PSNR of 38.85 and an exceptionally high SSIM of 0.995. These findings strongly suggest the successful simulation of the blurring process, indicating that our model performs exceptionally well in preserving image quality during deblurring.

\subsection{\textit{Pseudo-Inverse Simulation}}
\begin{figure}[!htb]
    \centering
    \includegraphics[width=0.49\textwidth]{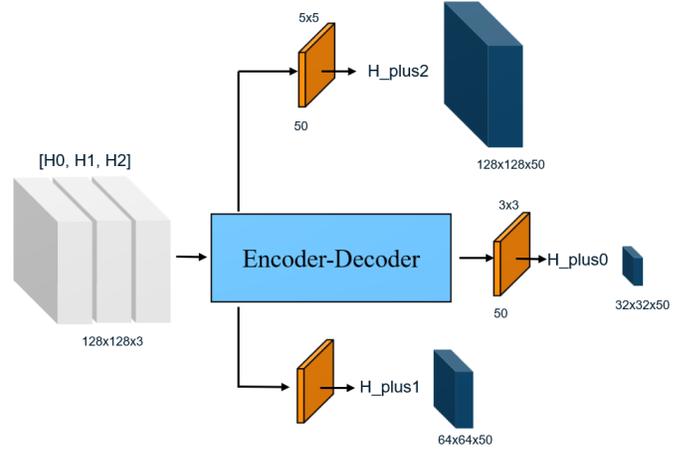}
    \caption{The network to calculate pseudo-inverse kernels $\bH^+$, which uses the same encoder and decoder structure with blurring simulation. At each upsampling level, the network outputs the corresponding blurring kernels $\bH^+_0$, $\bH^+_1$, and $\bH^+_2$.}
    \label{fig5}
\end{figure}

The simulation and computation of the pseudo-inverse process are also based on the similar encoder and decoder with Fig. \ref{fig2}. As shown in Fig. \ref{fig5}, the input consists of three results from the blurring process: $\bH_0$, $\bH_1$, and $\bH_2$. The sizes of $\bH_0$ and $\bH_1$ are adjusted to match $\bH_2$ through interpolation. After feature extraction and processing, corresponding pseudo-inverse kernels $\bH^+_0$, $\bH^+_1$, and $\bH^+_2$ are obtained.

It is worth noting that the input for the pseudo-inverse simulation network is the concatenation of three different dimensions of blurring simulation features, which are not directly related to $\by$ and $\bx$. In other words, to fit the pseudo-inverse process, a trained $\bH$ is required, which is why we separate the two steps. Additionally, the design of the loss function is another reason for this separation. 

Unlike blurring simulation, which directly uses the difference between $\by$ and $\bH\bx$ as the optimization target, pseudo-inverse simulation leverages the property $\bH\bH^+\bH\bx=\bH\bx$ and 

$\bH^+\bH\bH^+\bx=\bH^+\bx$. Similarly, we denote the Charbonnier loss function:
\begin{equation}
    Inverse\_Loss = \sum\limits_{i=0}^{2}\| \bH\bx_i - \bH\bH^+\bH\bx_i \|^2_2
    \label{eq:im_for5}
\end{equation}

Furthermore, unlike blurring simulation, which requires leveraging the ApplyH model only once, this part of the training process requires multiple iterations. Let me elaborate on this process in detail. Let's denote $\psi$ as the  ApplyH model. Then, we can proceed step by step to calculate the desired results.

First, compute $\bH^+\bH\bx$, also denoted as $\bH^+\by$:
    \begin{equation}
    \bH^+\bH\bx = \psi(\bH\bx, \bH^+_0, \bH^+_1, \bH^+_2),
    \label{eq:im_for6}
    \end{equation}
    Next, we can calculate $\bH\bH^+\bH\bx$, which serves as the primary component of the loss during training:
    \begin{equation}
    \bH\bH^+\bH\bx = \psi(\bH^+\bH\bx, \bH_0, \bH_1, \bH_2),
    \label{eq:im_for7}
    \end{equation}
    Similarly we can get an important result $\bH^+\bx$ format, which can be used in many restoration tasks:
    \begin{equation}
    \bH^+\bx = \psi(\bx, \bH^+_0, \bH^+_1, \bH^+_2),
    \label{eq:im_for8}
    \end{equation}
For example, with trained $\bH^+$, we can apply it to y to get $\bH^+\by$. To be more clear, the trained $\bH^+$ can be used on any input, including $\by$. Just for training process we use $\bx$. Thus we obtain the desired $\bH^+\by$, which could be a good estimation of the sharp. As shown in Table \ref{table2}, we evaluate the training effectiveness of this part of the network using numerical metrics.

\begin{table}[h]
\centering
    \caption{Metrics of Inverse Process. The PSNR, which measures image quality, is high, averaging at 59.69. SSIM values with a mean of 0.999.}
    \vskip 10pt
    \begin{tabular}{ccccccc}
    \hline \noalign{\smallskip}
    \multicolumn{1}{c}{} & \multicolumn{1}{c}{Valid1} & \multicolumn{1}{c}{Valid2} & \multicolumn{1}{c}{Valid3} & \multicolumn{1}{c}{Valid4} & \multicolumn{1}{c}{Valid5} & \multicolumn{1}{c}{Mean} \\ \noalign{\smallskip} \hline \noalign{\smallskip}
    Loss & 0.006       & 0.006       & 0.006       & 0.006       & 0.006       & 0.006   \\ \noalign{\smallskip}
    PSNR & 61.44       & 61.48       & 56.46       & 58.87       & 60.21       & 59.69   \\ \noalign{\smallskip}
    SSIM & 0.999       & 0.999       & 0.999       & 0.999       & 0.999       & 0.999   \\ \noalign{\smallskip} \hline
    \end{tabular}
    \label{table2}
\end{table}

The proposed method simulates blurring and pseudo-inverse processes using NAFNet blocks, achieving high PSNR and SSIM values, indicating successful restoration performance. Following this, we will introduce the Variational Deep Network to further enhance the simulation and restoration processes.

\subsection{\textit{Variational Deep-learning Model}}
First, we use the network from \cite{chu2022nafssr} as our baseline. The overall architecture is a UNet with a depth of 4. We use 28 NAFNet blocks concatenated for the main feature processing. Each of the encoder and decoder parts contains 4 NAFNet blocks, totaling 36 NAFNet blocks across four scales (1/1, 1/2, 1/4 and 1/8).

We adopt the deep variational framework (VDM) proposed in\cite{soh2022variational} to enhance performance. This framework conditions the restoration process using latent variables that incorporate domain and task-specific knowledge. These latent variables are estimated using a Variational Autoencoder (VAE), where we use NAFNet\cite{chu2022nafssr} as the component unit of the VAE. This approach of utilizing latent variables aligns with our idea. It is essential to note that DL-based models can solve inverse problems by learning the joint distribution $p(\bx, \by)$ that produces pairs of clean and degraded videos and estimating the posterior distribution $p(\bx|\by)$  from it.

\begin{figure}[!htb]
    \centering
    \includegraphics[width=0.49\textwidth]{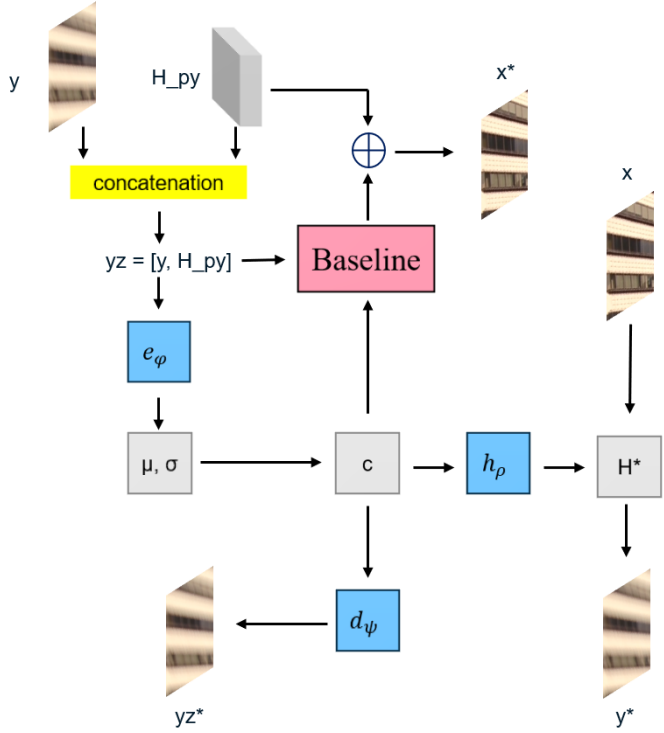}
    \caption{The network to deblur the video. The baseline is the same as \cite{chu2022nafssr}.}
    \label{fig6}
\end{figure}

 Let us introduce a new latent variable $\bc$ containing domain and task-specific information. Furthermore, let us assume that degraded videos are generated based on a procedure involving this variable 
$\bc$ in three steps:
\begin{enumerate}[nosep, itemsep=0.3em]
    \item  $\bc$ is generated using a domain prior $\mathcal{K}$ and a degradation process $T$ with $p(\bc;\mathcal{K},T)$. 
    \item The sharp $\bx$ is obtained by $\bc$ and an image prior $\mathcal{E}$ from $p(\bx|\mathcal{E};\bc)$
    \item The degraded $\by$ is from $p(\by|\bx,\bc)$
\end{enumerate}

This approximation allows us to model domain and taskspecific knowledge using $\bc$. However, it introduces the problem of estimating the posterior $p(\bc|\bx,\by)$ on top of $p(\bx|\by)$. From \cite{soh2022variational} we can find the solution by approximating the inference of $\log p(\bc|\bx,\by)$ and $\log p(\bx|\by)$ using the neural networks $\mathrm{b}_\theta$, $\mathrm{e}_\phi$ and $\mathrm{d}_\psi$:
    \begin{equation}
    \begin{aligned}
        \arg \max_{\theta, \phi, \psi} \; & \mathbb{E}_{p_{\text{data}}(\bx, \by)} \left[ \mathbb{E}_{\bc \sim q_{\phi}(\bc|\by)} \left[ \log p_{\theta}(\bx | \by, \bc) \right] \right] \\
        & - D_{\text{KL}} \left( q_{\phi}(\bx | \by) \| p(\bc) \right) \\
        & + \mathbb{E}_{\bc \sim q_{\phi}(\bc | \by)} \left[ \log p_{\psi}(\by | \bc) \right],
    \end{aligned}
    \label{eq:im_for9}
    \end{equation}
where $\theta$, $\phi$ and $\psi$ are the parameters of the three networks
b, e and d. $p_{\text{data}}(\bx, \by)$ means the underlying empirical data distribution, $D_{\text{KL}}$ is is the Kullback–Leibler divergence between two distributions. $\mathcal{L}_X$ and $\mathcal{L}_Y$ are all the Charbonnier loss. The first term in \eqref{eq:im_for9} could be approximated by a network $\mathrm{b}_\theta$. Here $\mathrm{b}_\theta$ is our baseline, thus it is input should be the concatenation of y, c and $\bH^+\by$ (\textbf{for simplicity, take the concatenation of $\by$ and $\bH^+\by$ as $\by\bz$}). This network is trained to minimize the distance between the real sharp video x and the predicted one $\bx^*$:
    \begin{equation}
       \mathcal{L}_1 = \mathcal{L}_X (\bx, b_{\theta} (\by\bz, e_{\phi} (\by\bz))) 
       \label{eq:im_for10}
    \end{equation}
    
where $e_{\phi} (\by\bz)$ is $\bc$. And $\mathrm{e}_\phi$ and $\mathrm{d}_\psi$ are encoder and decoder of VAE respectively. Based on the second term, we should calculate the loss between y and $\by^*$:
    \begin{equation}
        \mathcal{L}_2 = \lambda_{\text{rec}} \mathcal{L}_Y (\by\bz, d_{\psi} (e_{\phi} (\by)))
        \label{eq:im_for11}
    \end{equation}
Also, we need the loss:
    \begin{equation}
        \mathcal{L}_3 = \lambda_{\text{KL}} \text{KL}(e_{\phi}(\by\bz))
        \label{eq:im_for12}
    \end{equation}
Considering that we estimate $\bc$ during the training process, one additional model $\mathrm{h}_\rho$ and loss term should be introduced. The objective is to include features related to the degradation process into $\bc$. Thus $\mathrm{h}_\rho$ should be only used during training process.

Since the goal is to incorporate features related to the blurring process into 
$\bc$, we can directly utilize the blurring simulation network introduced in the earlier part of the paper. However, unlike the previous methods, the training of this part of the network does not require retaining the parameters of the blur kernels or the blurring model. Therefore, the parameters learned in each iteration can be used to directly degrade x, and only the degraded result $\by^*$ is kept for calculating the loss function. The specific process is illustrated in Fig. \ref{fig6}. Also,  the value range of these parameters is now comparable to $\bx$ and $\by$, making selecting the weights of the different losses easier. So, the extra loss could be:
    \begin{equation}
    \mathcal{L}_4 = \mathcal{L}_H (\by, h_{\rho}(e_{\phi}(\by\bz))_s * \bx_s)
    \label{eq:im_for13}
    \end{equation}
Then, we use $\lambda_{rec}$ and $\lambda_{vae}$ to balance $\mathcal{L}_1$ and the rest:
    \begin{equation}
    \mathcal{L}_{total} = \lambda_{rec}\mathcal{L}_1 + \lambda_{vae}(\mathcal{L}_2+\mathcal{L}_3+\mathcal{L}_4)
    \label{eq:im_for14}
    \end{equation}
Usually, people make the encoder more deep than the decoder, since the encoder is more important. But for our task, we have already included the pseudo-inverse estimation $\bH^+\by$ as a compensation, the encoder network here could be simpler. For encoder $\mathrm{e}_\phi$ and decoder $\mathrm{d}_\psi$, we use both 3 layers with 2, 2, 4 (or 4, 2, 2) totally 8 NAFNet blocks each side. $h\rho$ has the same structure as the decoder. 

%%huge ablation
\begin{table*}[ht]
\centering
    \caption{Comparisons of model size and FLOPS per pixel. FLOPS are calculated on an image with the resolution 128 x 128. The 'forward' indicates blurring simulation model. The 'backward' indicates pseudo-inverse simulation model.}
    \vskip 10pt
    \begin{tabular}{cccccccccc}
    \hline \noalign{\smallskip}
    & \multicolumn{1}{c}{EDVR\cite{b24}} & \multicolumn{1}{c}{TSP\cite{b28}} & \multicolumn{1}{c}{SRN\cite{b13}} & \multicolumn{1}{c}{DBN\cite{b15}} & \multicolumn{1}{c}{ESTRNN\cite{ESTRNN}} & \multicolumn{1}{c}{PVDNet\cite{Son2021PVDNet}} & \multicolumn{1}{c}{Ours} & \multicolumn{1}{c}{forward} & \multicolumn{1}{c}{backward} \\ 
    \noalign{\smallskip} \hline \noalign{\smallskip}
    Params(M) & 23.01 & 16.22 & 10.2 & 15.3 & 2.47 & 10.5 & 32.72 & 0.89 & 1.83 \\ 
    FLOPS(G) & 16.64 & 45.5 & 13.6 & 7.2 & 2.1 & 17.83 & 20.2 & 1.51 & 2.77 \\ 
    \noalign{\smallskip} \hline
    \end{tabular}
    \label{table3}
\end{table*}
\begin{figure*}[ht]
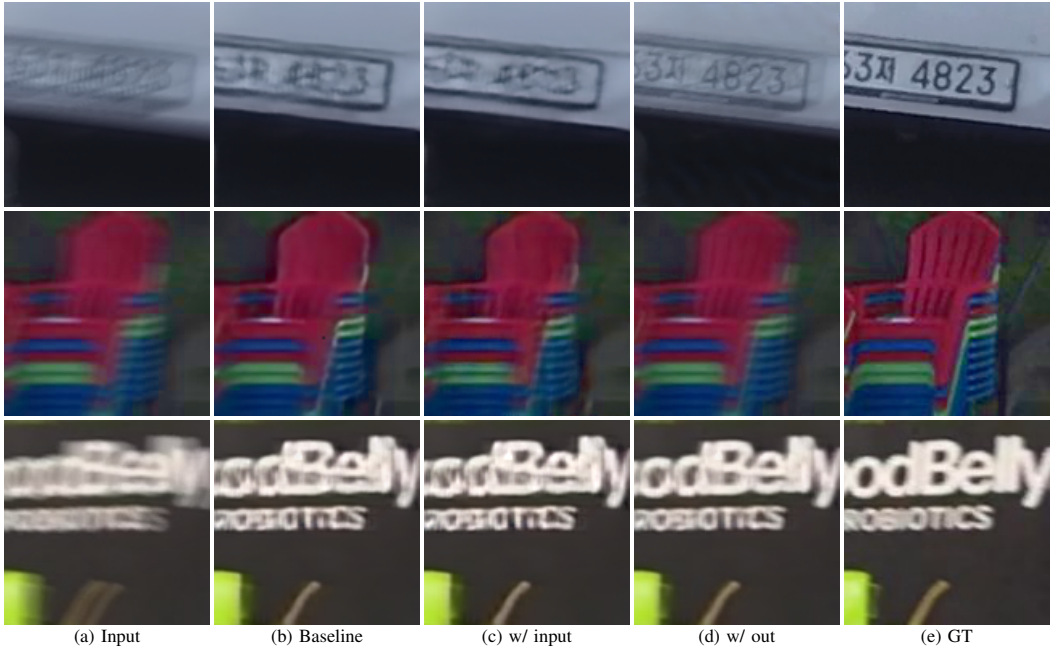

    \scriptsize  
    \renewcommand\arraystretch{0.7}
    \centering
    \begin{adjustbox}{width=\textwidth}
        \begin{minipage}[t]{\textwidth} % Full width for the grid of small images
            \centering
            \setlength{\tabcolsep}{1pt} % Adjust the spacing between images
            \begin{tabular}{ccccc}
                \includegraphics[width=0.15\textwidth]{debla/05_75/input.pdf} &
                \includegraphics[width=0.15\textwidth]{debla/05_75/base.pdf} &
                \includegraphics[width=0.15\textwidth]{debla/05_75/in.pdf} &
                \includegraphics[width=0.15\textwidth]{debla/05_75/Ours.pdf} &
                \includegraphics[width=0.15\textwidth]{debla/05_75/GT.pdf} \\
                
                \includegraphics[width=0.15\textwidth]{debla/29_41/input.pdf} &
                \includegraphics[width=0.15\textwidth]{debla/29_41/base.pdf} &
                \includegraphics[width=0.15\textwidth]{debla/29_41/in.pdf} &
                \includegraphics[width=0.15\textwidth]{debla/29_41/Ours.pdf} &
                \includegraphics[width=0.15\textwidth]{debla/29_41/GT.pdf} \\
                
                \includegraphics[width=0.15\textwidth]{debla/68_02/input.pdf} &
                \includegraphics[width=0.15\textwidth]{debla/68_02/base.pdf} &
                \includegraphics[width=0.15\textwidth]{debla/68_02/in.pdf} &
                \includegraphics[width=0.15\textwidth]{debla/68_02/Ours.pdf} &
                \includegraphics[width=0.15\textwidth]{debla/68_02/GT.pdf} \\
        
                (a) Input & (b) Baseline & (c) w/ input & (d) w/ out & (e) GT
            \end{tabular}
        \end{minipage}
    \end{adjustbox}
    \caption{Comparison of ablation study. Incorporating $\bH^+\by$ in the output section can restore more details and partially correct the distortion caused by the deblurring model.}
    \label{fig7}
\end{figure*}

%%single Gopro
\begin{figure*}[!htb]
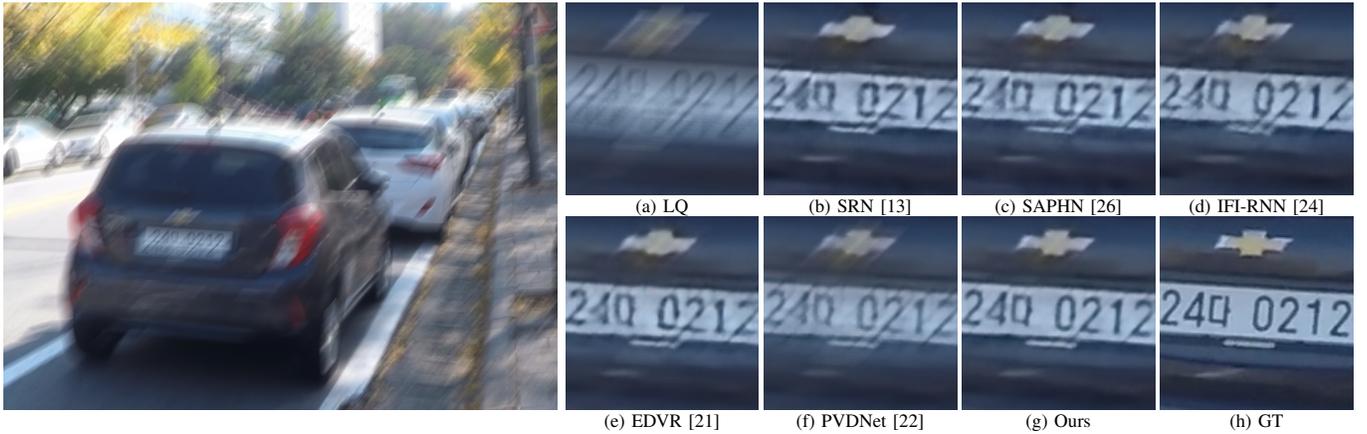

    \scriptsize  
    \renewcommand\arraystretch{0.7}
    \centering
    \begin{adjustbox}{width=\textwidth}
        \begin{minipage}[t]{0.4\textwidth} % Adjusted the width ratio for the large image
            \centering
            \raisebox{-0.463\height}{ % Adjust the height to align top edges
                \includegraphics[width=\textwidth,clip,trim=0 0 300 0]{Gopro/05_00/arigin.pdf}
            }
        \end{minipage}
        \hfill
        \begin{minipage}[t]{0.58\textwidth} % Adjusted the width ratio for the grid of small images
            \centering
            \setlength{\tabcolsep}{1pt} % Adjust the spacing between images
            \begin{tabular}{cccc}
                \includegraphics[width=0.24\textwidth]{Gopro/05_00/input.pdf} &
                \includegraphics[width=0.24\textwidth]{Gopro/05_00/SRN.pdf} &
                \includegraphics[width=0.24\textwidth]{Gopro/05_00/SAPHN.pdf} &
                \includegraphics[width=0.24\textwidth]{Gopro/05_00/IFIRNN.pdf} \\
                (a) LQ & (b) SRN \cite{b13} & (c) SAPHN \cite{SAPHN} & (d) IFI-RNN \cite{IFI-RNN} \\
                \includegraphics[width=0.24\textwidth]{Gopro/05_00/EDVR.pdf} &
                \includegraphics[width=0.24\textwidth]{Gopro/05_00/PVDNet.pdf} &
                \includegraphics[width=0.24\textwidth]{Gopro/05_00/Ours.pdf} &
                \includegraphics[width=0.24\textwidth]{Gopro/05_00/GT.pdf} \\
                (e) EDVR \cite{b24} & (f) PVDNet \cite{Son2021PVDNet} & (g) Ours & (h) GT
            \end{tabular}
        \end{minipage}
    \end{adjustbox}
    \caption{Comparison of different methods for video deblurring on GoPro dataset\cite{b38}. Our method, while achieving better overall deblurring, is also particularly effective in highlighting the strokes of the third Korean character.}
    \label{fig8}
\end{figure*}

\section{Experiment Settings}
\label{thrd:setting}
\subsection{\textit{Datasets}}

We use the following public datasets in our experiments:
\textbf{Gopro}\cite{b38} is composed of over $3000$ blurry-sharp image pairs of dynamic scenes captured by a high-speed camera. The dataset consists of $3214$ sequences have a resolution of $1280\times720$, and are divided into $2103$ training and $1111$ test sequences. It captures various real-world dynamic scenes under different conditions, including both indoor and outdoor environments. The training and testing subsets are split proportionally to 2:1. This dataset is particularly valuable for training and benchmarking deblurring algorithms due to its realistic blur effects created by averaging successive sharp frames from high-frame-rate video sequences.

\textbf{DVD}\cite{b39} dataset consists of $71$ videos with $6708$ blurry-sharp frame pairs, divided into train/test subsets with $61$ videos ($5708$ frame pairs) for training and $10$ videos ($1000$ frame pairs) for testing. The dataset is captured using mobile phones and DSLR cameras at a frame rate of $240$ fps. This provides a comprehensive set of dynamic scenes that include a variety of blurring effects caused by camera motion, making it suitable for training robust deblurring models.

%%Gopro
\begin{table*}[ht]
\centering
    \caption{Metrics of various video deblurring methods on Gopro dataset\cite{b38}. The PSNR, SSIM, STRRED, and LPIPS values indicate the performance of each method. Our method achieves the best PSNR and LPIPS, while SAPHN achieves the best SSIM. * Denotes the results reported in \cite{b49}}
    \vskip 10pt
    \begin{tabular}{cccccccccc}
    \hline \noalign{\smallskip}
    \multicolumn{1}{c}{} & \multicolumn{1}{c}{TSP*\cite{b28}} & \multicolumn{1}{c}{EDVR*\cite{b24}} & \multicolumn{1}{c}{ESTRNN*\cite{ESTRNN}} & \multicolumn{1}{c}{PVDNet\cite{Son2021PVDNet}} & \multicolumn{1}{c}{IFI-RNN\cite{IFI-RNN}} & \multicolumn{1}{c}{SFE\cite{b44}} & \multicolumn{1}{c}{SRN\cite{b13}} & \multicolumn{1}{c}{SAPHN\cite{SAPHN}} & \multicolumn{1}{c}{Ours} \\ \noalign{\smallskip} \hline \noalign{\smallskip}
    PSNR & 31.67       & 31.54       & 31.07       & 31.98       & 31.05       & 31.10       & 30.26       & 31.85       & \textbf{32.31} \\ \noalign{\smallskip}
    SSIM & 0.9279      & 0.9256      & 0.9023      & 0.9280      & 0.9110      & 0.9130      & 0.9342      & \textbf{0.9480}      & 0.9369 \\ \noalign{\smallskip}
    STRRED & 0.2153      & 0.2359      & 0.2688      & 0.1681      & 0.2647      & 0.2575      & 0.1992      & 0.1491      & \textbf{0.1283} \\ \noalign{\smallskip}
    LPIPS & 0.1061      & 0.1119      & 0.1289      & 0.0952      & 0.1297      & 0.1274      & 0.1593      & 0.0903      & \textbf{0.0845} \\ \noalign{\smallskip} \hline
    \end{tabular}
    \label{table4}
\end{table*}

\begin{figure*}[ht]
    \scriptsize  
    \renewcommand\arraystretch{0.7}
    \centering
    \begin{adjustbox}{width=\textwidth}
        \begin{minipage}[t]{0.4\textwidth} % Adjusted the width ratio for the large image
            \centering
            \raisebox{-0.463\height}{ % Adjust the height to align top edges
                \includegraphics[width=\textwidth,clip,trim=0 0 300 0]{Gopro/05_75/arigin.pdf}
            }
        \end{minipage}
        \hfill
        \begin{minipage}[t]{0.58\textwidth} % Adjusted the width ratio for the grid of small images
            \centering
            \setlength{\tabcolsep}{1pt} % Adjust the spacing between images
            \begin{tabular}{cccc}
                \includegraphics[width=0.24\textwidth]{Gopro/05_75/input.pdf} &
                \includegraphics[width=0.24\textwidth]{Gopro/05_75/SRN.pdf} &
                \includegraphics[width=0.24\textwidth]{Gopro/05_75/SAPHN.pdf} &
                \includegraphics[width=0.24\textwidth]{Gopro/05_75/IFIRNN.pdf} \\
                (a) LQ & (b) SRN \cite{b13} & (c) SAPHN \cite{SAPHN} & (d) IFI-RNN \cite{IFI-RNN} \\
                \includegraphics[width=0.24\textwidth]{Gopro/05_75/EDVR.pdf} &
                \includegraphics[width=0.24\textwidth]{Gopro/05_75/PVDNet.pdf} &
                \includegraphics[width=0.24\textwidth]{Gopro/05_75/Ours.pdf} &
                \includegraphics[width=0.24\textwidth]{Gopro/05_75/GT.pdf} \\
                (e) EDVR \cite{b24} & (f) PVDNet \cite{Son2021PVDNet} & (g) Ours & (h) GT
            \end{tabular}
        \end{minipage}
    \end{adjustbox}
    \label{fig:9upper}
\end{figure*}
\begin{figure*}[ht]
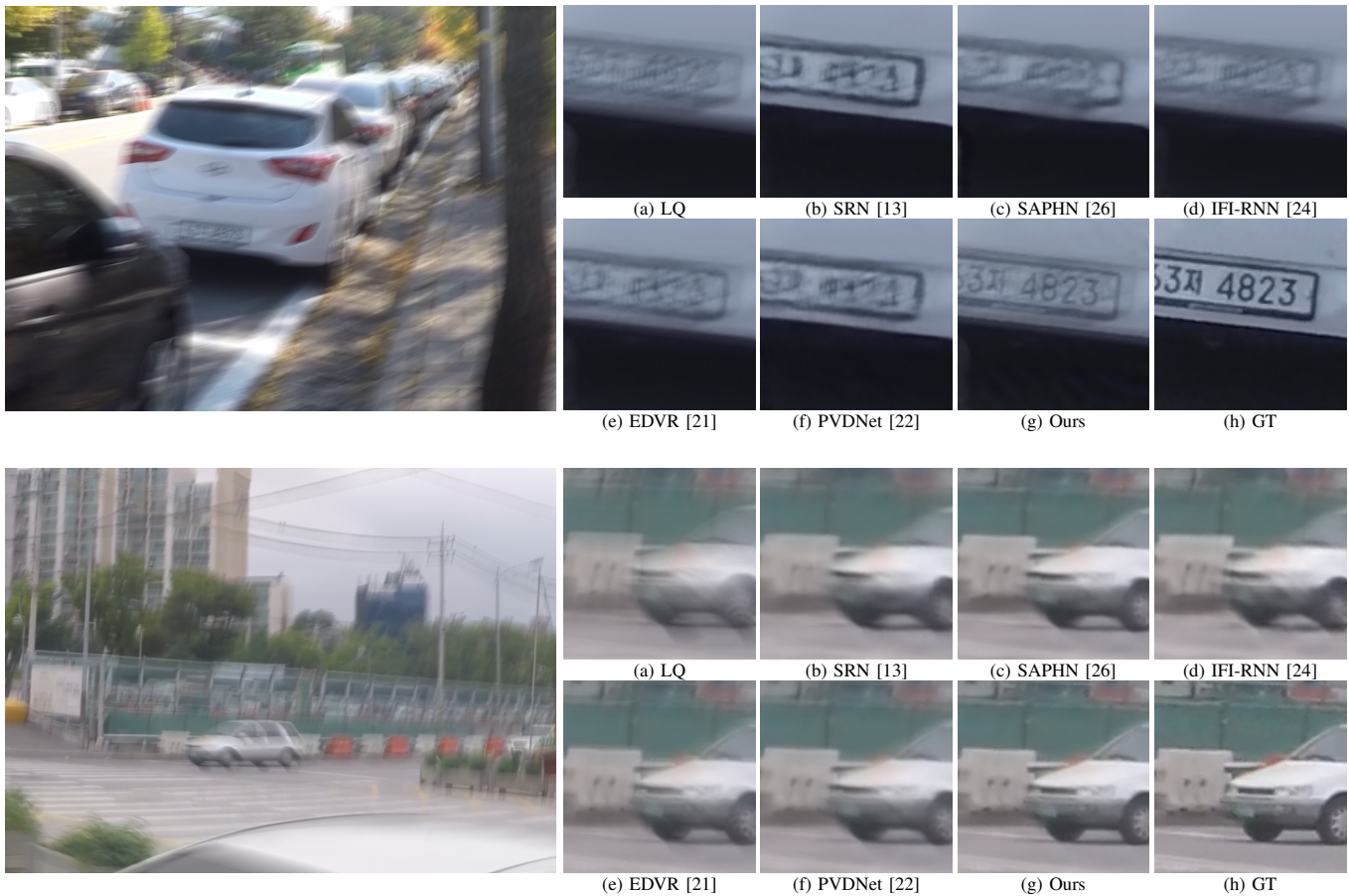

    \scriptsize  
    \renewcommand\arraystretch{0.7}
    \centering
    \begin{adjustbox}{width=\textwidth}
        \begin{minipage}[t]{0.4\textwidth} % Adjusted the width ratio for the large image
            \centering
            \raisebox{-0.463\height}{ % Adjust the height to align top edges
                \includegraphics[width=\textwidth,clip,trim=300 0 0 0]{Gopro/04_00/arigin.pdf}
            }
        \end{minipage}
        \hfill
        \begin{minipage}[t]{0.58\textwidth} % Adjusted the width ratio for the grid of small images
            \centering
            \setlength{\tabcolsep}{1pt} % Adjust the spacing between images
            \begin{tabular}{cccc}
                \includegraphics[width=0.24\textwidth]{Gopro/04_00/input.pdf} &
                \includegraphics[width=0.24\textwidth]{Gopro/04_00/SRN.pdf} &
                \includegraphics[width=0.24\textwidth]{Gopro/04_00/SAPHN.pdf} &
                \includegraphics[width=0.24\textwidth]{Gopro/04_00/IFIRNN.pdf} \\
                (a) LQ & (b) SRN \cite{b13} & (c) SAPHN \cite{SAPHN} & (d) IFI-RNN \cite{IFI-RNN} \\
                \includegraphics[width=0.24\textwidth]{Gopro/04_00/EDVR.pdf} &
                \includegraphics[width=0.24\textwidth]{Gopro/04_00/PVDNet.pdf} &
                \includegraphics[width=0.24\textwidth]{Gopro/04_00/Ours.pdf} &
                \includegraphics[width=0.24\textwidth]{Gopro/04_00/GT.pdf} \\
                (e) EDVR \cite{b24} & (f) PVDNet \cite{Son2021PVDNet} & (g) Ours & (h) GT
            \end{tabular}
        \end{minipage}
    \end{adjustbox}
    \caption{Comparison of different methods for video deblurring on GoPro dataset\cite{b38}. Our method is more robust in severe blurring scenario, like the upper part.}
    \label{fig9}
\end{figure*}

\textbf{REDS}\cite{Nah_2019_CVPR_Workshops_REDS} dataset, introduced in the NTIRE 2019 Challenge, includes $240$ training sequences, $30$ evaluation sequences, and $30$ testing sequences, each with $100$ frames. The video sequences have a resolution of $720\times1280$, capturing realistic and diverse scenes. REDS is designed to provide a benchmark for example-based video deblurring and super-resolution algorithms, making it a crucial resource for advancing these fields. It includes both indoor and outdoor scenes, and its high-quality frames are essential for developing and evaluating state-of-the-art image and video restoration techniques.

\subsection{\textit{Training details}}
We applied five types of image augmentations: random cropping to a size of $128\times128$, horizontal and vertical flipping, and random transposing. These increase the diversity of the training data. To train the model, we use the Adam optimizer\cite{Adam} with parameters $\beta_1=0.9$ and $\beta_2=0.999$. During training, the batch size is set to $32$ , and the learning rate is started at $10^{-3}$ and is gradually decayed to $10^{-6}$ using the cosine annealing schedule \cite{loshchilov2017sgdr}. The whole network is trained for $160$ epochs, with $5000$ iterations per epoch. The values for $\lambda_{rec}$ and $\lambda_{vae}$ are set to 1 and $5\times10^{-2}$, respectively, to balance the reconstruction loss and variational autoencoder loss. All computations were performed on a system with Intel(R) Xeon(R) CPU E5-2620 v4 @ 2.10GHz processor, 32GB of RAM, and a NVIDIA TITAN V GPU. The code is available at \href{https://github.com/zhihao0611/Video-deblur.git}{https://github.com/zhihao0611/Video-deblur.git}.

%%DVD
\begin{table*}[ht]
\centering
    \caption{Metrics of various video deblurring methods on DVD dataset\cite{b39}. Our method achieves the best PSNR and SSIM. * Denotes the results reported in \cite{b49}}
    \vskip 10pt
    \begin{tabular}{cccccccccc}
    \hline \noalign{\smallskip}
    \multicolumn{1}{c}{} & \multicolumn{1}{c}{TSP*\cite{b28}} & \multicolumn{1}{c}{EDVR*\cite{b24}} & \multicolumn{1}{c}{STFAN\cite{zhou2019stfan}} & \multicolumn{1}{c}{PVDNet\cite{Son2021PVDNet}} & \multicolumn{1}{c}{SFE\cite{b44}} & \multicolumn{1}{c}{ARVo*\cite{Li_2021_CVPR}} & \multicolumn{1}{c}{STTN\cite{Kim_2018_ECCV}} & \multicolumn{1}{c}{SRN\cite{b13}} & \multicolumn{1}{c}{Ours} \\ \noalign{\smallskip} \hline \noalign{\smallskip}
    PSNR & 32.13       & 31.82       & 31.15       & 32.31       & 31.71       & 32.80       & 31.61       & 30.53       & \textbf{32.95} \\ \noalign{\smallskip}
    SSIM & 0.9268      & 0.9160      & 0.9049      & 0.9260      & 0.9160      & 0.9352      & 0.9160      & 0.8940      & \textbf{0.9444} \\ \noalign{\smallskip}
    STRRED & 0.2221      & 0.2431      & 0.3159      & 0.1918      & 0.2614      & 0.1306      & 0.2709      & 0.3755      & \textbf{0.1159} \\ \noalign{\smallskip}
    LPIPS & 0.0998      & 0.1128      & 0.1375      & 0.0927      & 0.1177      & 0.0807      & 0.1196      & 0.1582      & \textbf{0.0759} \\ \noalign{\smallskip} \hline
    \end{tabular}
    \label{table5}
\end{table*}

\begin{figure*}[ht]
    \scriptsize  
    \renewcommand\arraystretch{0.7}
    \centering
    \begin{adjustbox}{width=\textwidth}
        \begin{minipage}[t]{0.4\textwidth} % Adjusted the width ratio for the large image
            \centering
            \raisebox{-0.463\height}{ % Adjust the height to align top edges
                \includegraphics[width=\textwidth,clip,trim=300 0 0 0]{DVD/37_35/arigin.pdf}
            }
        \end{minipage}
        \hfill
        \begin{minipage}[t]{0.58\textwidth} % Adjusted the width ratio for the grid of small images
            \centering
            \setlength{\tabcolsep}{1pt} % Adjust the spacing between images
            \begin{tabular}{cccc}
                \includegraphics[width=0.24\textwidth]{DVD/37_35/input.pdf} &
                \includegraphics[width=0.24\textwidth]{DVD/37_35/SRN.pdf} &
                \includegraphics[width=0.24\textwidth]{DVD/37_35/STFAN.pdf} &
                \includegraphics[width=0.24\textwidth]{DVD/37_35/TSP.pdf} \\
                (a) LQ & (b) SRN \cite{b13} & (c) STFAN \cite{zhou2019stfan} & (d) TSP \cite{b28} \\
                \includegraphics[width=0.24\textwidth]{DVD/37_35/EDVR.pdf} &
                \includegraphics[width=0.24\textwidth]{DVD/37_35/PVDNet.pdf} &
                \includegraphics[width=0.24\textwidth]{DVD/37_35/Ours.pdf} &
                \includegraphics[width=0.24\textwidth]{DVD/37_35/GT.pdf} \\
                (e) EDVR \cite{b24} & (f) PVDNet \cite{Son2021PVDNet} & (g) Ours & (h) GT
            \end{tabular}
        \end{minipage}
    \end{adjustbox}
    % \caption{Comparison of different methods for video deblurring on DVD dataset.}
    \label{fig10upper}
\end{figure*}
\begin{figure*}[ht]
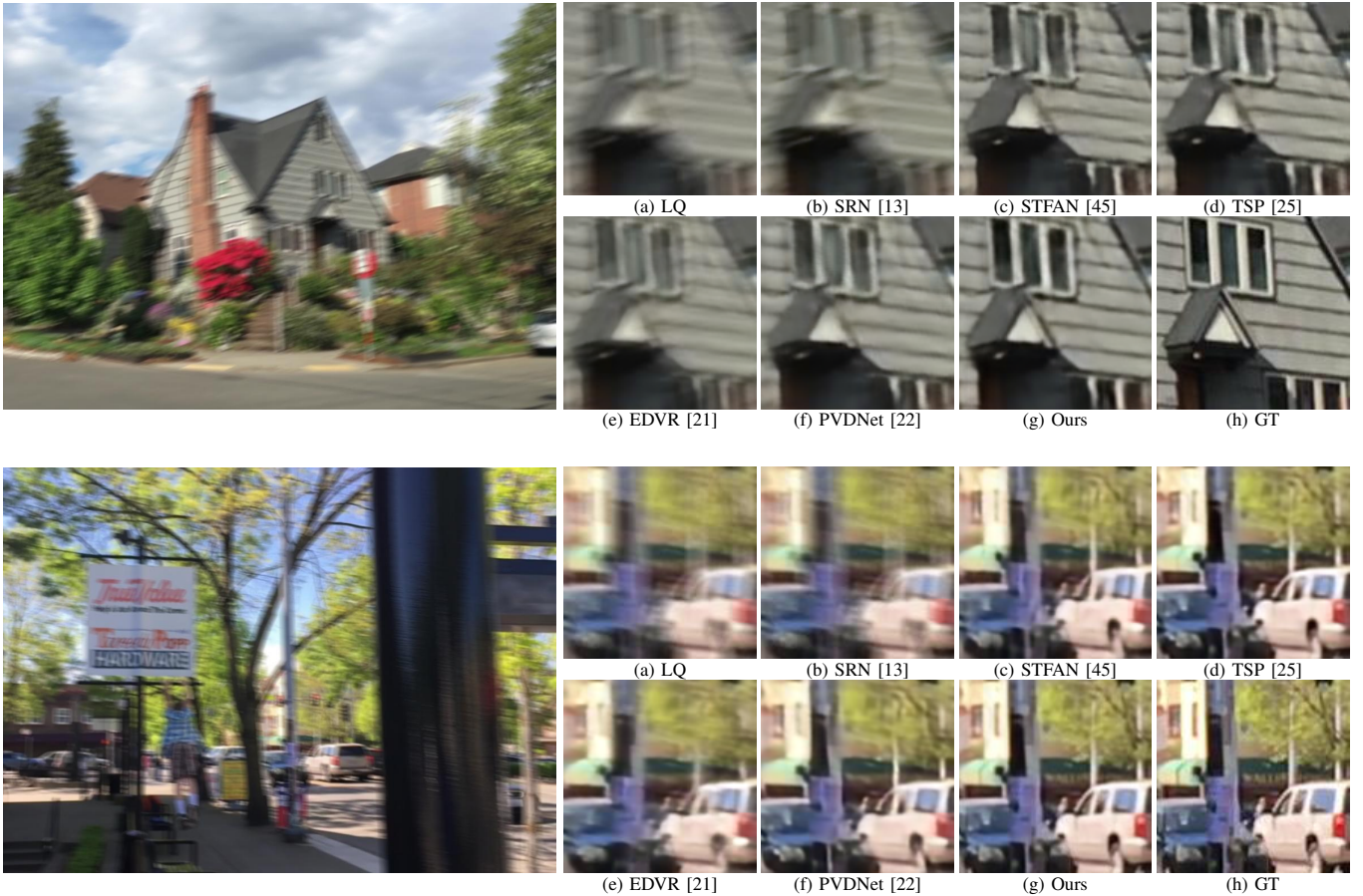

    \scriptsize  
    \renewcommand\arraystretch{0.7}
    \centering
    \begin{adjustbox}{width=\textwidth}
        \begin{minipage}[t]{0.4\textwidth} % Adjusted the width ratio for the large image
            \centering
            \raisebox{-0.463\height}{ % Adjust the height to align top edges
                \includegraphics[width=\textwidth,clip,trim=300 0 0 0]{DVD/29_41/arigin.pdf}
            }
        \end{minipage}
        \hfill
        \begin{minipage}[t]{0.58\textwidth} % Adjusted the width ratio for the grid of small images
            \centering
            \setlength{\tabcolsep}{1pt} % Adjust the spacing between images
            \begin{tabular}{cccc}
                \includegraphics[width=0.24\textwidth]{DVD/29_41/input.pdf} &
                \includegraphics[width=0.24\textwidth]{DVD/29_41/SRN.pdf} &
                \includegraphics[width=0.24\textwidth]{DVD/29_41/STFAN.pdf} &
                \includegraphics[width=0.24\textwidth]{DVD/29_41/TSP.pdf} \\
                (a) LQ & (b) SRN \cite{b13} & (c) STFAN \cite{zhou2019stfan} & (d) TSP \cite{b28} \\
                \includegraphics[width=0.24\textwidth]{DVD/29_41/EDVR.pdf} &
                \includegraphics[width=0.24\textwidth]{DVD/29_41/PVDNet.pdf} &
                \includegraphics[width=0.24\textwidth]{DVD/29_41/Ours.pdf} &
                \includegraphics[width=0.24\textwidth]{DVD/29_41/GT.pdf} \\
                (e) EDVR \cite{b24} & (f) PVDNet \cite{Son2021PVDNet} & (g) Ours & (h) GT
            \end{tabular}
        \end{minipage}
    \end{adjustbox}
    \caption{Comparison of different methods for video deblurring on DVD dataset\cite{b39}. For the upper part, our method better removes the overlapping blur caused by the window frames. For the lower part, we restore the text on the green facade of the restaurant better.}
    \label{fig10}
\end{figure*}

\section{Ablation Study}
\label{four:ablation}

% The core of our method are introducing $\mathrm{H^+y}$ and variational deep-learning model.
We experiment by progressively adding components of our proposed model to the baseline to quantify its contribution to the overall performance. The following models are considered in the ablation study:
\begin{enumerate}[nosep, itemsep=0.3em]
    \item baseline: The same architecture as the one proposed in \cite{chu2022nafssr} with $36$ NAFNet blocks.
    \item w/ input: Concatenating y and $\bH^+\by$ at the input of the baseline 
    \item w/ output: Continuing adding $\bH^+\by$ at the end of the model.
    \item w/VDN: Continuing adding $\mathrm{e}_\phi$ and $\mathrm{d}_\psi$ to the model. But without $\mathcal{L}_4$ and $\mathrm{h}_\rho$ during training.
    \item Our complete model.
\end{enumerate}

As shown in Table~\ref{table6}, on the DVD dataset, compared to the baseline, introducing $\bH^+\by$ at the input alone significantly improves PSNR by $1.15$ dB and SSIM by $0.0120$. Adding $\bH^+\by$ at the output as well further improves PSNR by $0.05$ dB and SSIM by $0.0009$. Adding $\mathrm{e}_\phi$ and $\mathrm{d}_\psi$ increase PSNR $0.08$ dB, SSIM $0.0004$. Adding $\mathrm{h}_\rho$ increases PSNR $0.04$ dB, SSIM $0.0003$, totalling the use of VDN to and increase of $0.12$ dB PSNR, $0.0007$ SSIM.

On the GoPro dataset we observe a similar situation, with concatenating $\bH^+\by$ on input increasing PSNR by $1.07$ dB and SSIM by $0.0145$. Adding $\bH+^\by$ further improves PSNR by $0.06$ dB and SSIM by $0.0007$, with our complete solution (using VDN with $\mathcal{L}_4$ and $\mathrm{h}_\rho$) managing to push the performance further in terms of PSNR and SSIM by $0.11$ dB and $0.0011$, respectively.

Finally, on REDS dataset we see a very similar scenario, where each proposed component pushing further the performing, till the different between our complete model and the baseline is of $1.41$ dB and $0.0206$ PSNR and SSIM, respectively.

\begin{table*}[ht]
\centering
    \caption{Metrics of various video deblurring methods on REDS dataset\cite{Nah_2019_CVPR_Workshops_REDS}. Our method achieves the best PSNR and SSIM. * Denotes the results reported in \cite{ESTRNN}}
    \vskip 10pt
    \begin{tabular}{ccccccc}
    \hline \noalign{\smallskip}
    \multicolumn{1}{c}{} & \multicolumn{1}{c}{ESTRNN*\cite{ESTRNN}} & \multicolumn{1}{c}{STRCNN\cite{b18}} & \multicolumn{1}{c}{IFI-RNN\cite{IFI-RNN}} & \multicolumn{1}{c}{DBN\cite{b15}} & \multicolumn{1}{c}{Ours} \\ \noalign{\smallskip} \hline \noalign{\smallskip}
    PSNR & 32.63       & 30.23       & 31.36       & 31.55       & \textbf{32.91} \\ \noalign{\smallskip}
    SSIM & 0.9110      & 0.8708      & 0.8942      & 0.8960      & \textbf{0.9262} \\ \noalign{\smallskip}
    STRRED & 0.1674      & 0.4071      & 0.2901      & 0.2727      & \textbf{0.1334} \\ \noalign{\smallskip}
    LPIPS & 0.0971      & 0.1699      & 0.1377      & 0.1294      & \textbf{0.0874} \\ \noalign{\smallskip} \hline
    \end{tabular}
    \label{table7}
\end{table*}

\begin{figure*}[ht]
    \scriptsize  
    \renewcommand\arraystretch{0.7}
    \centering
    \begin{adjustbox}{width=\textwidth}
        \begin{minipage}[t]{0.54\textwidth} % Adjusted the width ratio for the large image
            \centering
            \raisebox{-0.47\height}{ % Adjust the height to align top edges
                \includegraphics[width=\textwidth,clip,trim=300 0 0 0]{REDS/00_09/arigin.pdf}
            }
        \end{minipage}
        \hfill
        \begin{minipage}[t]{0.6\textwidth} % Adjusted the width ratio for the grid of small images
            \centering
            \setlength{\tabcolsep}{1pt} % Adjust the spacing between images
            \begin{tabular}{ccc}
                \includegraphics[width=0.32\textwidth]{REDS/00_09/input.pdf} &
                \includegraphics[width=0.32\textwidth]{REDS/00_09/DBN.pdf} &
                \includegraphics[width=0.32\textwidth]{REDS/00_09/IFIRNN.pdf} \\
                (a) LQ & (b) DBN \cite{b15} & (c) IFI-RNN \cite{IFI-RNN} \\
                \includegraphics[width=0.32\textwidth]{REDS/00_09/ESTRCNN.pdf} &
                \includegraphics[width=0.32\textwidth]{REDS/00_09/Ours.pdf} &
                \includegraphics[width=0.32\textwidth]{REDS/00_09/GT.pdf} \\
                (d) ESTRNN \cite{ESTRNN} & (e) Ours & (f) GT
            \end{tabular}
        \end{minipage}
    \end{adjustbox}
    \label{fig11upper}
\end{figure*}
\begin{figure*}[ht]
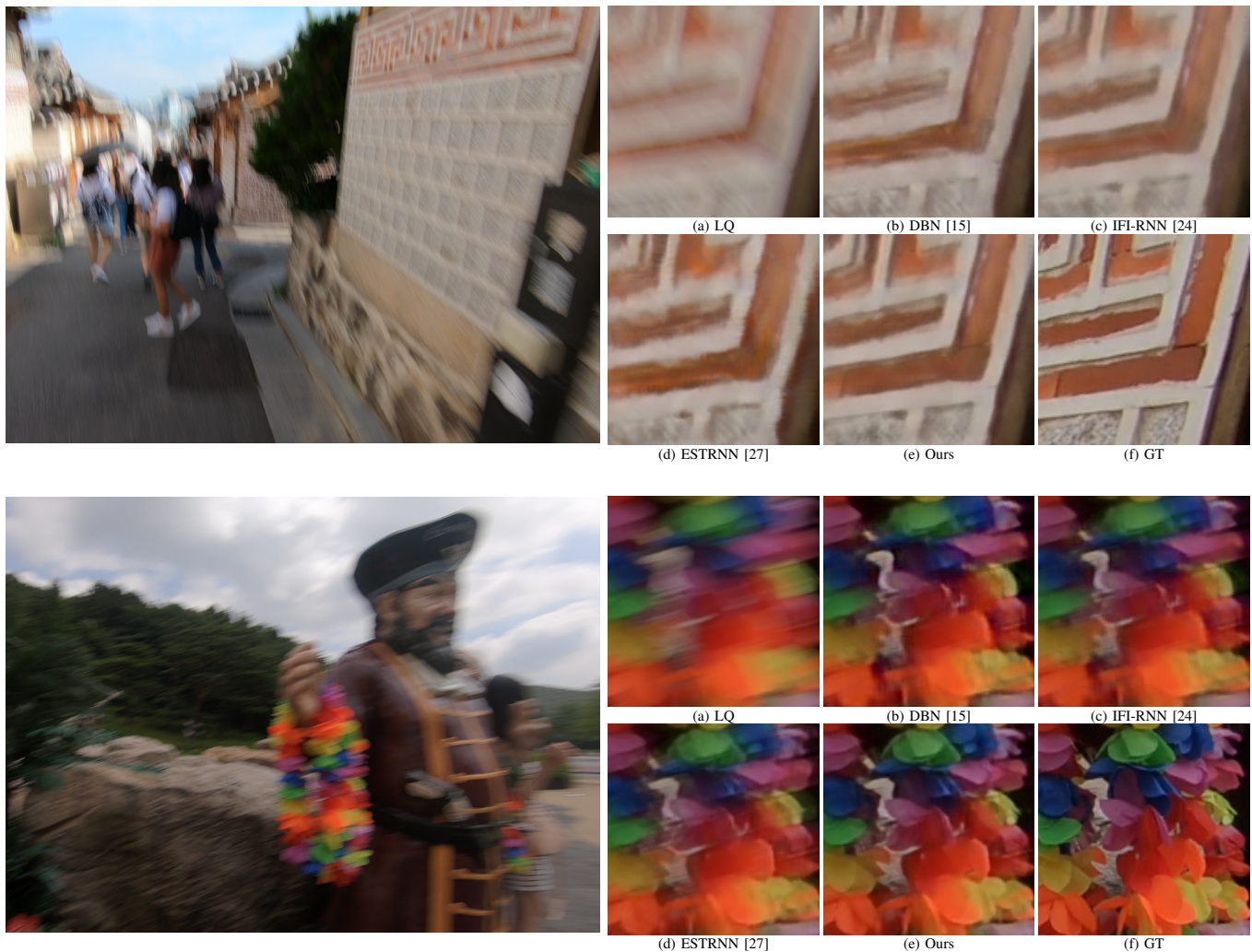

    \scriptsize  
    \renewcommand\arraystretch{0.7}
    \centering
    \begin{adjustbox}{width=\textwidth}
        \begin{minipage}[t]{0.54\textwidth} % Adjusted the width ratio for the large image
            \centering
            \raisebox{-0.47\height}{ % Adjust the height to align top edges
                \includegraphics[width=\textwidth,clip,trim=300 0 0 0]{REDS/06_60/arigin.pdf}
            }
        \end{minipage}
        \hfill
        \begin{minipage}[t]{0.6\textwidth} % Adjusted the width ratio for the grid of small images
            \centering
            \setlength{\tabcolsep}{1pt} % Adjust the spacing between images
            \begin{tabular}{ccc}
                \includegraphics[width=0.32\textwidth]{REDS/06_60/input.pdf} &
                \includegraphics[width=0.32\textwidth]{REDS/06_60/DBN.pdf} &
                \includegraphics[width=0.32\textwidth]{REDS/06_60/IFIRNN.pdf} \\
                (a) LQ & (b) DBN \cite{b15} & (c) IFI-RNN \cite{IFI-RNN} \\
                \includegraphics[width=0.32\textwidth]{REDS/06_60/ESTRCNN.pdf} &
                \includegraphics[width=0.32\textwidth]{REDS/06_60/Ours.pdf} &
                \includegraphics[width=0.32\textwidth]{REDS/06_60/GT.pdf} \\
                (d) ESTRNN \cite{ESTRNN} & (e) Ours & (f) GT
            \end{tabular}
        \end{minipage}
    \end{adjustbox}
    \caption{Comparison of different methods for video deblurring on REDS dataset\cite{Nah_2019_CVPR_Workshops_REDS}. Our method better preserves the contour information of the patterns for the upper scenario.}
    \label{fig11}
\end{figure*}

% \begin{table}[ht]
% \centering
%     \caption{Ablation study on DVD, Gopro, and REDS datasets. The PSNR and SSIM values indicate the performance of each method.}
%     \vskip 10pt
%     \begin{tabular}{ccccccc}
%     \hline \noalign{\smallskip}
%     & \multicolumn{2}{c}{DVD} & \multicolumn{2}{c}{Gopro} & \multicolumn{2}{c}{REDS} \\ \noalign{\smallskip} \hline \noalign{\smallskip}
%     & PSNR & SSIM & PSNR & SSIM & PSNR & SSIM \\ \noalign{\smallskip} \hline \noalign{\smallskip}
%     baseline & 31.63 & 0.9308 & 31.07 & 0.9206 & 31.50 & 0.9056 \\ \noalign{\smallskip}
%     w/ input & 32.78 & 0.9391 & 32.14 & 0.9333 & 32.70 & 0.9221 \\ \noalign{\smallskip}
%     w/ out & 32.83 & 0.9410 & 32.20 & 0.9340 & 32.79 & 0.9231 \\ \noalign{\smallskip}
%     w/ VDN & 32.91 & 0.9413 & 32.27 & 0.9344 & 32.88 & 0.9236 \\ \noalign{\smallskip}
%     Ours & 32.95 & 0.9415 & 32.30 & 0.9347 & 32.91 & 0.9239 \\ \noalign{\smallskip} \hline
%     \end{tabular}
%     \label{table6}
% \end{table}
\begin{table}[ht]
\centering
    \caption{Ablation study on DVD, Gopro, and REDS datasets. The PSNR, SSIM, STRRED, and LPIPS values indicate the performance of each method.}
    \vskip 10pt
    \begin{tabular}{ccccccc}
    \hline \noalign{\smallskip}
    & \multicolumn{1}{c}{} & \multicolumn{1}{c}{baseline} & \multicolumn{1}{c}{w/ input} & \multicolumn{1}{c}{w/ out} & \multicolumn{1}{c}{w/ DVN} & \multicolumn{1}{c}{Ours} \\ \noalign{\smallskip} \hline \noalign{\smallskip}
    DVD & PSNR & 31.63 & 32.78 & 32.83 & 32.91 & \textbf{32.95} \\ \noalign{\smallskip}
        & SSIM & 0.9308 & 0.9428 & 0.9437 & 0.9441 & \textbf{0.9444} \\ \noalign{\smallskip}
        & STRRED & 0.2593 & 0.1345 & 0.1285 & 0.1209 & \textbf{0.1159} \\ \noalign{\smallskip}
        & LPIPS & 0.0902 & 0.0816 & 0.0799 & 0.0772 & \textbf{0.0759} \\ \noalign{\smallskip} \hline \noalign{\smallskip}
    Gopro & PSNR & 31.07 & 32.14 & 32.20 & 32.27 & \textbf{32.31} \\ \noalign{\smallskip}
          & SSIM & 0.9206 & 0.9351 & 0.9358 & 0.9364 & \textbf{0.9369} \\ \noalign{\smallskip}
          & STRRED & 0.2617 & 0.1445 & 0.1385 & 0.1319 & \textbf{0.1283} \\ \noalign{\smallskip}
          & LPIPS & 0.0943 & 0.0896 & 0.0878 & 0.0856 & \textbf{0.0845} \\ \noalign{\smallskip} \hline \noalign{\smallskip}
    REDS & PSNR & 31.50 & 32.70 & 32.79 & 32.88 & \textbf{32.91} \\ \noalign{\smallskip}
         & SSIM & 0.9056 & 0.9242 & 0.9251 & 0.9257 & \textbf{0.9262} \\ \noalign{\smallskip}
         & STRRED & 0.2843 & 0.1538 & 0.1449 & 0.1371 & \textbf{0.1334} \\ \noalign{\smallskip}
         & LPIPS & 0.1036 & 0.0041 & 0.0912 & 0.0886 & \textbf{0.0874} \\ \noalign{\smallskip} \hline
    \end{tabular}
    \label{table6}
\end{table}

%%REDS

\begin{figure}[!htb]
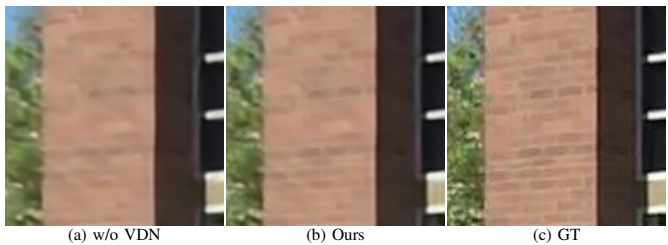
 % Use 'H' to force placement
    \renewcommand\arraystretch{0.7}
    \centering
    \begin{adjustbox}{width=0.49\textwidth} % Adjust the width to half of the page
        \setlength{\tabcolsep}{1pt} % Adjust the spacing between images
        \begin{tabular}{ccc}
            \includegraphics[width=0.3\textwidth]{debla/16_52/wo.pdf} &
            \includegraphics[width=0.3\textwidth]{debla/16_52/Ours.pdf} &
            \includegraphics[width=0.3\textwidth]{debla/16_52/GT.pdf} \\
            \large (a) w/o VDN & \large (b) Ours & \large (c) GT
        \end{tabular}
    \end{adjustbox}
    \caption{Comparison of ablation study. Incorporating VDN can achieve better vision quality.}
    \label{fig12}
\end{figure}

The numerical evaluations show that concatenating $\bH^+\by$ at the input significantly improves performance, while adding it to the output features yields smaller gains. However, the qualitative results in Fig. \ref{fig7} reveal that adding $\bH^+\by$ at the output effectively enhances the restoration of contours and details in heavily blurred scenes. Introducing $\bH^+\by$ at both input and output better restores details in extreme blur scenarios. Moreover, characters often exhibit distortion after deblurring under such extreme conditions, and introducing $\bH^+\by$ at the output can reduce this distortion to some extent. Therefore, introducing $\bH^+\by$ at the output is necessary. Moreover, introducing the VDN helps to improve the visual quality of the images. As shown in Fig. \ref{fig12}, after adding the VDN, the details of the wall tiles are clearer.

Because our network incorporates two pre-trained networks, we are concerned not only with deblurring performance but also with the network's parameter count and operational cost. As shown in Table \ref{table3}, our network has a parameter count of $32.72$M. It is important to note that the parameter counts for the blurring simulation model and the pseudo-inverse simulation model are only $0.89$M and $1.83$M, respectively. FLOPS are $1.51$G and $2.77$G.

%%small ablation

This means that the total parameter count of our final network is largely determined by the size of the baseline. Although our baseline model\cite{chu2022nafssr} has a large parameter count, its optimization of computational complexity results in relatively low FLOPS.

\section{Comparison With State Of The Art.}
\label{five:comparison}

We adopt public available source codes for evaluation. For some methods that do not have results on some datasets in their original paper, we have retrained these models to calculate the metrics with the same training strategy described in their original papers, and code provided by the authors.

We first analyze the results on the GoPro\cite{b38} dataset. As shown in Table \ref{table4}, our method achieves the highest PSNR value, outperforming the second-best PVDNet\cite{Son2021PVDNet} by $0.33$ dB. Moreover, we also achieve the best perfomance on LPIPS ($0.0845$) and STRRED ($0.1283$), showing that our model achieves at the same time the best perceptual quality and time consistency. While SAPHN\cite{SAPHN} shows better performance in terms of SSIM (+$0.111$), our method  outperforms it in every other metric, excelling in handling severe blurring, as demonstrated by the qualitative comparison in Fig. \ref{fig8}. The challenge in this scenario is that the third character is a Korean letter, which appears more like the digit '0' or the uppercase letter 'Q' in the blurred image. Among the compared methods, PVDNet\cite{Son2021PVDNet}, IFI-RNN\cite{IFI-RNN}, and our method can distinguish the original Korean character's strokes from those of '0' and 'Q'. However, PVDNet\cite{Son2021PVDNet} is less effective in removing motion blur, and IFI-RNN\cite{IFI-RNN} introduces some distortion. Our method shows superior performance in both maintaining the character's distinctive strokes and effectively reducing camera motion blur. This ability to preserve fine details while eliminating motion blur highlights the effectiveness of our method in practical deblurring tasks. To better demonstrate our method's superior handling of extreme camera motion blur, we conducted tests on more challenging scenarios.

Fig. \ref{fig9} further demonstrates the robustness of our method in handling severe camera motion blur. In the severe blurring scenario shown in the figure, our method is able to preserve a relatively complete contour of the intruder, while other methods exhibit varying degrees of distortion. This indicates the superior performance of our method in mitigating camera-induced motion blur. Additionally, we are concerned with object motion blur, which results from the movement of objects themselves. The bottom part of Fig. \ref{fig9} illustrates such a scenario, where our method excels in restoring the fast running car, demonstrating its effectiveness in addressing object-induced blur.

To further validate the robustness of our method across different scenes and types of blur, we tested it on additional datasets. The numerical results on the DVD dataset\cite{b39} are presented in Table \ref{table5}. Our method achieved the highest performance across all metrics, outperforming the second-best method by $0.15$ dB in PSNR, $0.0092$ in SSIM, $0.0147$ in STRRED and $0.0048$ in LPIPS. In the top part of Fig. \ref{fig10}, we showcase the deblurring effect on objects with overlapping blur, where the blur primarily affects the windows. Our method effectively removes the overlapping blur caused by window frames and glass. The bottom part of Fig. \ref{fig10} illustrates the restoration of text in really small size, where our method better preserves the original details of the text. These results collectively highlight the robustness and effectiveness of our approach in various challenging scenarios.

Finally, we present the results obtained on the REDS dataset\cite{Nah_2019_CVPR_Workshops_REDS}. As shown in Table \ref{table7}, our method again shows the best perfomance across all metrics, outperforming the second-best method by $0.28$ dB in PSNR, $0.0152$ in SSIM, $0.034$ STRRED and $0.0097$ LPIPS. Thus, our model excels at perceptual quality and temporal consistency when compared against the state-of-the-art in REDS. Additionally, Fig. \ref{fig11} demonstrates that our method produces clearer contours and edges in the deblurred results (e.g., the boundary of the wall brick patterns). This indicates that our method also performs remarkably well on this dataset.

Based on our extensive experiments across the GoPro\cite{b38}, DVD\cite{b39}, and REDS\cite{Nah_2019_CVPR_Workshops_REDS} datasets, our method consistently demonstrates superior performance in both quantitative and qualitative metrics.

\section{Conclusions}
\label{six:conclusion}

In this paper, we introduced VDPI, a novel video deblurring method that combines blur pseudo-inverse modeling with deep learning. Our key innovation is using CNNs to fit both the blurring process and its pseudo-inverse, enhancing the model's adaptability and performance.

Our approach was tested on the GoPro\cite{b38}, DVD\cite{b39}, and REDS\cite{Nah_2019_CVPR_Workshops_REDS} datasets, consistently achieving superior results compared to state-of-the-art methods. The inclusion of the blur pseudo-inverse estimation $\bH^+\by$ at both the input and output stages and the use of a Variational Deep Network (VDN) proved crucial in restoring fine details and contours in heavily blurred videos , while achieving a temporally consisting output.

Future work will explore further optimizations and adaptations of our method to broader applications in video processing and restoration tasks. One possibility is to optimize the training process and loss formulation of the unified blurring simulation model and pseudo-inverse simulation model, treating these two steps as an end-to-end network training process. In the same way, it is also possible to apply the pseudo-inverse $\bH^+\by$ method together with more advanced networks such as transformers and diffusion models.

\bibliographystyle{IEEEtran}
\bibliography{biblio}

% \begin{IEEEbiography}[{\includegraphics[width=1in,height=1.25in,clip,keepaspectratio]{zhihao.pdf}}]{Your Name}
% Biography text here. Describe your academic background, research interests, and other relevant details.
% \end{IEEEbiography}

% \begin{IEEEbiography}[{\includegraphics[width=1in,height=1.25in,clip,keepaspectratio]{santi.pdf}}]{Co-author's Name}
% Biography text here. Describe your co-author's academic background, research interests, and other relevant details.
% \end{IEEEbiography}

% \begin{IEEEbiography}[{\includegraphics[width=1in,height=1.25in,clip,keepaspectratio]{akk.pdf}}]{Co-author's Name}
% Biography text here. Describe your co-author's academic background, research interests, and other relevant details.
% \end{IEEEbiography}

\end{document}